\documentclass{article}
\usepackage{ijcai20}

\usepackage{times}

\usepackage{soul}
\usepackage{url}
\usepackage[hidelinks]{hyperref}
\usepackage[utf8]{inputenc}
\usepackage[small]{caption}
\usepackage{graphicx}
\usepackage{amsmath}
\usepackage{amsthm}
\theoremstyle{definition}
\newtheorem{definition}{Definition}[section]
\newtheorem{proposition}{Proposition}[section]
\newtheorem{theorem}{Theorem}[section]

\DeclareMathOperator*{\argmax}{arg\,max}
\DeclareMathOperator*{\argmin}{arg\,min}
\usepackage{amsfonts,amssymb}
\usepackage{bbm}
\usepackage{dsfont}
\usepackage{bm}
\usepackage[ruled,linesnumbered]{algorithm2e}
\usepackage{booktabs}
\usepackage{multirow}
\usepackage{subfigure}
\usepackage{microtype}
\usepackage[toc,page]{appendix}
\usepackage{pgfplots}
\pgfplotsset{compat=newest}
\usetikzlibrary{plotmarks}

\newcommand{\proofsketch}{\vspace*{-1ex} \noindent {\textit{Proof Sketch: }}}

\usepackage{xspace}
\makeatletter
\DeclareRobustCommand\onedot{\futurelet\@let@token\@onedot}
\def\@onedot{\ifx\@let@token.\else.\null\fi\xspace}

\def\ie{\emph{i.e}\onedot}

\def\etal{\emph{et al}\onedot}
\makeatother

\title{Towards Accurate and Robust Domain Adaptation under Noisy Environments}

\author{
Zhongyi Han$^1$\and
Xian-Jin Gui$^2$\and
Chaoran Cui$^{3*}$\And
Yilong Yin$^1$\footnote{Co-corresponding author}\\
\affiliations
$^1$School of Software, Shandong University, Jinan 250101. China\\
$^2$National Key Laboratory for Novel Software Technology, Nanjing University, Nanjing 210023, China\\
$^3$School of Computer Science and Technology, Shandong University of Finance and Economics\\
\emails
hanzhongyicn@gmail.com,
guixj@lamda.nju.edu.cn,
crcui@sdufe.edu.cn,
ylyin@sdu.edu.cn
}

\begin{document}

\maketitle

\begin{abstract}
In non-stationary environments, learning machines usually confront the domain adaptation scenario where the data distribution does change over time. Previous domain adaptation works have achieved great success in theory and practice. However, they always lose robustness in noisy environments where the labels and features of examples from the source domain become corrupted. In this paper, we report our attempt towards achieving accurate noise-robust domain adaptation. We first give a theoretical analysis that reveals how harmful noises influence unsupervised domain adaptation. To eliminate the effect of label noise, we propose an offline curriculum learning for minimizing a newly-defined empirical source risk. To reduce the impact of feature noise, we propose a proxy distribution based margin discrepancy. We seamlessly transform our methods into an adversarial network that performs efficient joint optimization for them, successfully mitigating the negative influence from both data corruption and distribution shift. A series of empirical studies show that our algorithm remarkably outperforms state of the art, over 10\% accuracy improvements in some domain adaptation tasks under noisy environments.

\end{abstract}

\section{Introduction}

Conventional learning theories assume that the learning machines are under a static environment where we draw training and test examples from an identical distribution. If the training and test distributions substantially differ, the learning machines would probably lose generalization ability~\cite{valiant1984theory}. However, we expect our learning machines can perform well in crucial non-stationary environments where the test domain is similar yet distinct from the training domain. Unsupervised domain adaptation is the key machine learning topic to deal with this problem~\cite{ben2007analysis}.

Prominent theoretical advance and effective algorithm have been achieved in domain adaptation. Ben-David \etal \shortcite{ben2007analysis,ben2010theory} propose a pioneering $\mathcal{H} \Delta \mathcal{H}$ distance and give rigorous generalization bounds to measure the distribution shift. A series of seminal studies then extend the $\mathcal{H} \Delta \mathcal{H}$ distance to discrepancy distance~\cite{mansour2009domain}, to domain disagreement distance~\cite{germain2013pac}, to generalized discrepancy distance~\cite{cortes2019adaptation}, or to margin disparity discrepancy~\cite{DBLP:conf/icml/0002LLJ19}, etc. With significant theoretical findings, domain adaptation algorithms have made significant advances. Previous studies have explored various techniques concerning statistic moment matching-based algorithms~\cite{pan2010domain,long2017deep}, gradual transition-based algorithms~\cite{gopalan2011domain}, and pseudo labeling-based algorithms~\cite{sener2016learning,saito2017asymmetric}. More interestingly, adversarial learning-based algorithms introducing a domain discriminator for minimizing distribution discrepancy, yields state of the art performance on many visual tasks~\cite{DBLP:conf/icml/GaninL15,tzeng2017adversarial,long2018conditional}.



While previous works have achieved significant successes, they easily fail in realistic scenarios. The reason is that they implicitly assume an ideal learning environment where the source data are noise-free, which is difficult to hold in practice. \textit{Domain adaptation in noisy environments} relaxes the assumption of clean data in standard domain adaptation to more realistic scenarios, where labels and features of examples from source domain are noisy. However, this more general yet challenging problem is under-explored so far. There is only one pioneering work~\cite{DBLP:conf/aaai/ShuCLW19} that proposes an online curriculum learning algorithm to handle this problem. Although this work makes some progress, how to conduct theoretical analysis and construct a robust algorithm for unsupervised domain adaptation under noisy environments is still an open problem.

In this paper, we give a theoretical analysis of unsupervised domain adaptation when trained with noisy labels and features. Our theoretical analysis reveals that label noise worsens the expected risk of the source distribution; thus, we define a conditional-weighted empirical risk and propose an offline curriculum learning, which also provides some effective remedial measures for the online curriculum learning~\cite{DBLP:conf/aaai/ShuCLW19}. Meanwhile, our theoretical analysis reveals that feature noise mainly aggravates the distribution discrepancy between source and target distribution; thus, we propose a novel proxy margin discrepancy by introducing a proxy distribution to provide an optimal solution for distribution discrepancy minimization. We finally transform our methods into an adversarial network that performs dynamic joint optimization between them. A series of empirical studies on synthetic and real datasets show that our algorithm remarkably outperforms previous methods.


\section{Domain Adaptation in Noisy Environments} 
\subsection{Learning Set-Up}
\label{setup}

In domain adaptation, a sample of $m$ labeled training examples $\{(x_i, y_i)\}_{i=1}^m$ is drawn according to a \textit{source distribution} $Q$ defined on $\mathcal{X} \times \mathcal{Y}$, where $\mathcal{X}$ is the feature set and $\mathcal{Y}$ is the label set. $\mathcal{Y}$ is $\{1, \dots, K\}$ in multi-class classification. Meanwhile, a sample of $n$ unlabeled test examples $\{(x_i)\}_{i=1}^n$ is drawn according to a \textit{target distribution} $P$ that somewhat differs from $Q$. Assume that the distributions $Q$ and $P$ should not be dissimilar substantially~\cite{DBLP:journals/jmlr/Ben-DavidLLP10}.

For domain adaptation in noisy environments, we relax the assumption of clean data in standard domain adaptation to that source distribution may be corrupted with feature noise and label noise independently. The source distribution is corrupted into a \textit{noisy source distribution} $Q_n$. We denote by $\tilde{x}$ the noisy features, and $p_f$ the probability of corrupting one feature with harmful noises $e$, \ie, $p(\tilde{x}_i\!=\!x_i\!+\!e)\!=\!p_f$ and $p(\tilde{x}_i\!=\!x_i)\!=\!1\!-\!p_f$. We denote by $\tilde{y}$ the noise labels corrupted with a noise transition matrix $\mathcal{T} \in \mathbb{R}^{K\times K}$ where $\mathcal{T}_{ij} = p(\tilde{y}=j|y=i)$ denotes the probability of labeling an $i$-th class example as $j$.

We denote by $L: \mathcal{Y} \times \mathcal{Y} \rightarrow \mathbb{R}$ a loss function defined over pairs of labels. For multi-class classification, we denote by $f: \mathcal{X} \rightarrow \mathbb{R}^K$ a scoring function, which induces a labeling function $h_f: \mathcal{X} \rightarrow \mathcal{Y}$ where $h_f: x \rightarrow \argmax_{y\in \mathcal{Y}} [f(x)]_y$. For any distribution $Q$ on $\mathcal{X} \times \mathcal{Y}$ and any labeling function $h\in \mathcal{H}$, we denote $\epsilon_{Q}(h) = \mathbb{E}_{(x,y)\sim Q} L(h(x),y)$ the expected risk. Our objective is to select a hypothesis $f$ out of a hypothesis set $\mathcal{F}$ with a small expected risk $\epsilon_{P}(h_f)$ on the target distribution.




\subsection{Theoretical Analysis}
\label{analysis}
For in-depth analysis, we derive an upper bound of expected target risk in noisy environments according to the triangle inequality. Full proofs will be provided in the longer version.

\begin{proposition} 
For any hypothesis $h \in \mathcal{H}$, the bound of target expected risk $\epsilon_{P}(h_f)$ in noisy environments is given by
\begin{equation}
    \epsilon_{P}(h) \le \epsilon_{Q_n}(h) + |\epsilon_{Q_n}(h, h^*) - \epsilon_{P}(h, h^*)| + \lambda \,,
    \label{error bound}
\end{equation}
where $\lambda \!=\! \epsilon_{Q_n}(h^*)\! + \!\epsilon_{P}(h^*)$ is the ideal combined error of $h^*$:
\begin{equation}
    h^* = \argmin_{h\in \mathcal{H}} \epsilon_{P}(h) + \epsilon_{Q_n}(h)\,.
\end{equation}
\label{proposition 1}
\end{proposition}

\proofsketch 
\textit{According to the triangle inquality, we have}
\begin{equation}
\begin{split}
    \epsilon_{P}(h) &\le \epsilon_{P}(h^*) + \epsilon_{P}(h, h^*)\\
    &= \epsilon_{P}(h^*) + \epsilon_{Q_n}(h, h^*) + \epsilon_{P}(h, h^*) - \epsilon_{Q_n}(h, h^*)\\
    &\le \epsilon_{P}(h^*) + \epsilon_{Q_n}(h, h^*) + |\epsilon_{P}(h, h^*) - \epsilon_{Q_n}(h, h^*)|\\
    &\le \epsilon_{Q_n}(h) + |\epsilon_{Q_n}(h, h^*) - \epsilon_{P}(h, h^*)| \\ 
    & \quad + [ \epsilon_{Q_n}(h^*) + \epsilon_P(h^*)]\\
\end{split}
\end{equation}

This bound is informative if the three terms are small enough; however, noises do worsen all of them as follows.

\textit{The first term}, expected risk of the noisy source distribution, is given by $\epsilon_{Q_n}(h) = \mathbb{E}_{(\tilde{x},\tilde{y})\sim Q_n} L(h(\tilde{x}),\tilde{y})$. Label noise worsens the generalization ability of $h$ in the first term, and the more label noises, the worse generalization ability. However, it is uncertain how feature noise behaves. Ilyas \etal \shortcite{ilyas2019adversarial} concluded that deep neural networks~(DNN) trained with adversarial perturbations might have better generalization and transfer-ability. Our empirical studies also find little difference between reserving and eliminating feature-noise examples when minimizing empirical source risk using DNNs. 

\textit{The second term} represents the measure of distribution discrepancy between the noisy source distribution and target distribution. Feature noise is dependent of it, but label noise is not. The following theorem reveals the dependence.

\begin{theorem} Assume the source distribution $Q$ be corrupted by feature noise into a new distribution $R$ while target distribution $P$ is unaffected. Assume feature noise is harmful and independent of the source and target distributions. Feature noise will enlarge the distribution discrepancy (disc) between source and target distribution, \ie \ $disc(Q, P) \le disc(R, P)$.
\label{th2.2}
\end{theorem}

\proofsketch \textit{Given distributions $Q_X$ and $P_X$ over feature space $\mathcal{X}$, assuming $\mathcal{H}$ denotes a symmetric hypothesis class, the $\mathcal{H}$-divergence~\cite{ben2007analysis} is derived into
\begin{equation}
\resizebox{.98\linewidth}{!}{$
    \displaystyle
    d_{\mathcal{H}}(Q,P) =2\Big(1-\inf_{h\in \mathcal{H}}\big(\Pr_{x\sim Q_X}[h(x)=0]+ \Pr_{x\sim P_X}[h(x)=1]\big)\Big)\,.
$}
\end{equation}%
When injecting harmful feature noise into source examples, the infimum would become lower, thus $d_{\mathcal{H}}(Q,P) \le d_{\mathcal{H}}(R,P)$.}

\textit{The third term} $\lambda$ is assumed to be small enough~\cite{DBLP:journals/jmlr/Ben-DavidLLP10}; however, if feature-noise and label-noise are heavy, the ideal hypothesis error of target distribution will be unbounded, and this bound would be uninformative.

In summary, the above analysis suggests two critical aspects of achieving robust domain adaptation. Firstly, we should achieve a \textit{robust empirical risk minimization} over the noisy source distribution by mitigating the negative influence of label noise. Secondly, we should concern a \textit{robust distribution discrepancy} by reducing the impact of feature noise.

\section{The Proposed Methods}
Based on our analysis, we present the offline curriculum learning for the robust empirical risk minimization~(Sec.~\ref{ERM}), and then introduce the robust proxy margin discrepancy~(Sec.~\ref{disc}).

\subsection{Offline Curriculum Learning}
\label{ERM}

To improve the empirical source risk, a natural idea is to eliminate label noise examples to recover a clean distribution. The pioneering work~\cite{DBLP:conf/aaai/ShuCLW19} adopts an online curriculum learning that defines a loss value threshold $\gamma$ to eliminate noisy examples in each training epoch $t\in T$ by
\begin{align}
     & \min_{f} \frac{1}{m} \sum_{i=1}^m \bm{w}_i^t L (f(\tilde{x}_i;\theta_f^t), \tilde{y}_i)\, , \, \text{where} \\
     & \bm{w}_{i}^t = \mathds{1}(\ell_{i}^t \le \gamma), \ i = 1, \dots, m, \ t = 1, \dots, T\,,
\end{align}
where $\bm{w}_i^t$ indicates whether or not to keep the $i$-th example, and $\theta_f^t$ is the parameter of hypothesis $f$ in the $t$-th epoch. $\ell_{i}^t = L(f(\tilde{x}_i;\theta_f^t), \tilde{y}_i)$ minimized by stochastic gradient descent (SGD). Symbol $\mathds{1}$ is the indicator function. $\gamma$ is set according to cross-validation. The reason behind its success is the \textit{small loss} criterion that reveals clean examples have smaller loss values than noisy examples~\cite{DBLP:conf/nips/HanYYNXHTS18}.

While online curriculum learning is intuitive, there are still four serious issues. \textit{Firstly}, when source data are class-imbalanced, the magnitude of loss values will exist substantial differences across different classes; online curriculum learning will cause biased estimation. \textit{Secondly}, it is in the expectation that correct examples will have smaller losses than incorrect ones in every epoch. However, the SGD algorithm makes the loss value of each example fluctuate in different epochs, which would cause the online curriculum learning unstable and unreliable to some extent. \textit{Thirdly}, setting a fixed threshold of $\gamma$ by cross-validation, is unpractical in domain adaptation. We cannot access the annotated data of the target domain. It is also unreasonable because it will lead DNNs to overfit on label-noise examples when the loss values of label-noise examples drop gradually with the increasing of the epoch. \textit{Finally}, online curriculum learning treats feature and label noise examples indiscriminately and removing all of them in the whole course. However, eliminating feature-noise examples is needless when minimizing empirical source risk according to our analysis because lots of valuable features would be wasted in such a way. To resolve the above issues, we define the robust conditional-weighted empirical risk.


\theoremstyle{definition}
\begin{definition}[\textbf{Conditional-Weighted Empirical Risk}]
Denote by $T$ the epoch number to filter out noisy examples, and denote by $\gamma_k$ the loss value threshold of class $k\in K$, we define the conditional-weighted empirical risk over noisy source distribution for any classifier $f\in \mathcal{F}$ by
\begin{align}
    \label{risk}
     & \epsilon_{\hat{Q}_n}(f) = \frac{1}{m} \sum_{i=1}^m w(\tilde{x}_i, \tilde{y}_i) L (f(\tilde{x}_i), \tilde{y}_i)\,, \, \text{where} \\
     \label{weight}
     & w(\tilde{x}_i, \tilde{y}_i) = \mathds{1}(\bar{\ell}_{i} \le \gamma_{\tilde{y}_i}), \ i = 1, \dots, m,\ \text{and}\\
     \label{loss mean}
     & \bar{\ell}_{i} = \frac{1}{T} \sum_{t=1}^T L (f(\tilde{x}_i; \theta_f^t), \tilde{y}_i)\,.
\end{align}
\label{def1}
\end{definition}

To optimize this risk, we propose the offline curriculum learning, which consists of two steps. In the first step, we construct the early training curriculum to filter out noisy examples. In short, we first learn a classifier $f$ and collect examples' loss values in $T$ epochs. We then average each example's loss value (in Eq.~\eqref{loss mean}) and rank them by class in ascending order. We use the loss value of the $(m_k\!\times\!p_k)$-th example as the loss value threshold $\gamma_k$ of class $k$. $m_k$ is the example number of class $k$, and $p_k = 1-r_k$ in which $r_k$ denotes the label-noise rate of class $k$. In practice, we find it is safer to set $p_k = \max\{1\!-\!1.2r_k, 0.8(1\!-\!r_k)\}$. Finally, we conduct an offline exam by comparing each example's averaged loss value of $\bar{\ell}_i$ with $\gamma_{\tilde{y}_i}$ to decide each example's weight $w(\tilde{x}_i, \tilde{y}_i)$.

In the second step, we minimize Eq.~\eqref{risk} using the trusted examples ($w(\tilde{x}_i, \tilde{y}_i)\!=\!1$) and use them to do domain adaptation. We choose cross-entropy loss, which can be optimized by SGD efficiently. Denote by $\sigma$ the softmax function, \ie, for $z \in \mathbb{R}^K$, $\sigma_k(z) \triangleq \frac{e^{z_k}}{\sum_{k=1}^K e^{z_k}}$, Eq.~\eqref{risk} is optimized by
\begin{equation}
    \min_{f} - \mathbb{E}_{(\tilde{x}, \tilde{y})\sim \hat{Q}_n} w(\tilde{x}_i,\tilde{y}_i)\log[ \sigma_{\tilde{y}}(f(\tilde{x}))]\,.
\label{loss_min_1}
\end{equation}

Offline curriculum learning (OCL) provides point-to-point remedial measures for online curriculum learning. Firstly, OCL ranks and selects small-loss examples class by class, fully considering the class-conditional information and successfully achieving an unbiased estimation. Secondly, OCL examines the average loss value of each example along the whole curriculum process for comparison, avoiding the randomness of loss values in different epochs. Thirdly, OCL gets an efficient threshold of $\gamma$ by leveraging the estimated label-noise rate, which can be estimated by existing methods~\cite{DBLP:conf/cvpr/PatriniRMNQ17}. We argue that we cannot expect to obtain a reliable classifier behind learning guarantees without known rough noise information. Finally, OCL only mitigates the negative influence of label noise but keeps valuable information on feature-noise examples to get better generalization. One might wonder how to separate label-noise and feature-noise examples. Our empirical studies point out that in the training process of DNNs with SGD, as for examples with the same noisy label, the ones with feature noises only will have smaller losses than the ones with false labels. Furthermore, the selected trusted source data by OCL directly supports shrinking the ideal combined hypothesis error of $\lambda$; such establishes a valid learning guarantee for the expected target risk.

\subsection{Proxy Margin Discrepancy}
\label{disc}
Recall that the measure of distribution discrepancy appeared as a critical term in the target error bound; thus, we especially propose the proxy margin discrepancy to mitigate feature noise.

Theorem~\ref{th2.2} implies that a better guarantee would hold if we could select, instead of $Q_n$, other \textit{proxy distribution} $Q'$ with a smaller discrepancy $disc(Q', P)$. We thus introduce the proxy discrepancy using the $Q'$ to give an optimal solution for distribution discrepancy minimization, by shrinking the discrepancy between source, proxy, and target distributions.


\begin{definition}[\textbf{Proxy Discrepancy, PD}]
Given a proxy distribution $Q'$, a noisy source distribution $Q_n$, and a target distribution $P$, for any measure of distribution discrepancy $disc$, we define the Proxy Discrepancy and its empirical version by
\begin{equation}
    \begin{split}
        &d_{\Lambda}(Q_n, P) = disc(Q_n, Q') + disc(Q', P)\,,\\
        &d_{\Lambda}(\hat{Q}_n, \hat{P}) = disc(\hat{Q}_n, \hat{Q}') + disc(\hat{Q}', \hat{P})\,.
    \end{split}
\end{equation}
\end{definition}

In practice, we need to select an efficient $disc$.  In the seminal works, Zhang \etal \shortcite{DBLP:conf/icml/0002LLJ19} propose the novel margin disparity discrepancy (MDD) that embeds margin loss into a disparity discrepancy with informative generalization bound.

\begin{definition}[Margin Disparity Discrepancy, MDD]
Given a hypothesis set $\mathcal{F}$, and a specific classifier $f\in \mathcal{F}$, the MDD and its empirical version induced by $f'\in \mathcal{F}$ are defined by
\begin{equation}
\begin{split}
    &d_{f, \mathcal{F}}^{(\rho)}(Q, P) = \sup_{f'\in \mathcal{F}} \Big (disp_P^{(\rho)}(f', f) - disp_Q^{(\rho)}(f', f)\Big) \,, \\
    &d_{f, \mathcal{F}}^{(\rho)}(\hat{Q}, \hat{P}) = \sup_{f'\in \mathcal{F}} \Big (disp_{\hat{P}}^{(\rho)}(f', f) - disp_{\hat{Q}}^{(\rho)}(f', f)\Big) \,,
\end{split}
\end{equation}
where $disp^{(\rho)}(f, f')$ is the margin disparity defined below. Let $\rho_{f'}$ denote the margin of a hypothesis $f'$ and $\Phi_{\rho}$ denote $\rho$-margin loss, the margin disparity is defined by
\begin{equation}
    disp_Q^{(\rho)}(f', f) = \mathbb{E}_{x \sim Q} \Phi_{\rho}(\rho_{f'}(x, h_f(x)))\,.
\label{margin_disparity}
\end{equation}
\label{MDD}
\end{definition}

\begin{definition}[\textbf{Proxy Margin Discrepancy, PMD}]
With the definition of MDD and a proxy distribution $Q'$, we define the Proxy Margin Discrepancy and its empirical version by
\begin{equation}
    \begin{split}
        &d_{\Lambda}^{(\rho)} (Q_n, P)  =  d_{f, \mathcal{F}}^{(\rho)} (Q_n, Q') + d_{f, \mathcal{F}}^{(\rho)} (Q', P) \,,\\
        &d_{\Lambda}^{(\rho)} (\hat{Q}_n, \hat{P})  =  d_{f, \mathcal{F}}^{(\rho)} (\hat{Q}_n, \hat{Q}') + d_{f, \mathcal{F}}^{(\rho)} (\hat{Q}', \hat{P}) \,.\\
    \end{split}
\end{equation}
\end{definition}
To reveal the capacity of proxy distribution, we derive a generalization bound for PMD based on the bound of MDD.

\begin{figure}[t]
\centering
\includegraphics[width=0.48\textwidth]{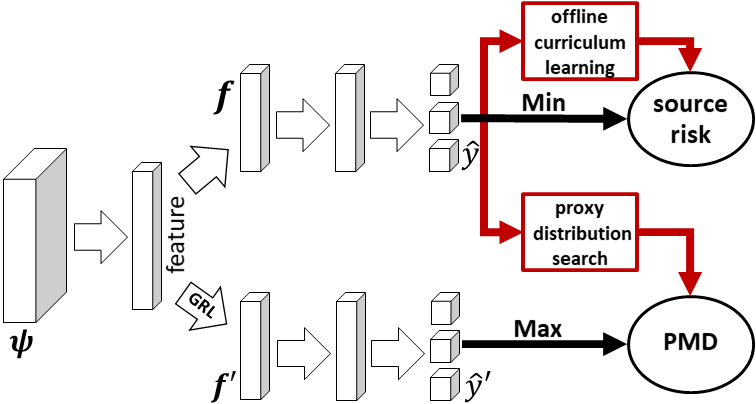}
\caption{The adversarial network for robust domain adaptation.}
\label{adversarial}
\end{figure}

\begin{theorem}
Based on the definition of Rademacher Complexity~\cite{mansour2009domain}, for any $\delta > 0$, with probability $1\!-\!2\delta$ over a sample of size $u$ drawn from $Q'$, a sample of size $m$ drawn from $Q$, and a sample of size $n$ dram from $P$, the following holds simultaneously for any scoring function $f$:
\begin{equation}
    \begin{split}
        & |d_{\Lambda}^{(\rho)} (\hat{Q}_n, \hat{P}) - d_{\Lambda}^{(\rho)} (Q_n, P)|  \\
        &\quad \le \frac{2K}{\rho} \mathfrak{R}_{m,Q_n} (\Pi_H^\mathcal{F}) + \frac{4K}{\rho} \mathfrak{R}_{u,Q'}(\Pi_H^\mathcal{F}) + \frac{2K}{\rho} \mathfrak{R}_{n,P} (\Pi_H^\mathcal{F}) \\
        &\quad+\sqrt{\frac{\log \frac{2}{\delta}}{2m}} +  2\sqrt{\frac{\log \frac{2}{\delta}}{2u}} + \sqrt{\frac{\log \frac{2}{\delta}}{2n}} \,.
    \end{split}
\end{equation}
where $\mathfrak{R}$ denote Rademacher complexity, $\Pi_H^\mathcal{F}$ is defined as $\{x \rightarrow f(x,h(x))|h\in H, f\in \mathcal{F}\}$, where $h$ is induced by $f$.
\label{lem2.1}
\end{theorem}

To achieve effective optimization of PMD, we adopt a deep adversarial learning network using cross-entropy loss instead of the margin loss of $\Phi_{\rho}$ according to~\cite{DBLP:conf/icml/0002LLJ19}. As shown in Fig.~\ref{adversarial}, it consists of a representation function $\psi$ and two classifiers $f$ and $f'$ with coincide to Definition~\ref{MDD}. Theorem~\ref{lem2.1} implies searching for an empirical proxy distribution with large size of $u$ is crucial. To achieve that, we assume that $\hat{Q}'$ is a better distribution that can be derived from noisy source distribution $\hat{Q}_n$, and let classifier $f$ search for it by
\begin{equation}
    \hat{Q}' = \argmin_{\hat{Q}' \in \mathcal{Q}} - \mathbb{E}_{{(\tilde{x}, y)\sim \hat{Q}'}}  \log [\sigma_{y}(f(\psi(\tilde{x})))]\,,
    \label{searchQ}
\end{equation}
where $\mathcal{Q}$ denotes the set of distributions with support of trusted source examples by $supp(\bm{w}\hat{Q}_n)$. When the size of $\hat{Q}'$ is zero, optimizing Eq.~\eqref{searchQ} is meaningless. Thus we limit the size of $\hat{Q}'$ by $|\hat{Q}'| \ge \tau |\bm{w}\hat{Q}_n|$ where $\tau \in [0,1]$ controls the ratio between $\hat{Q}'$ and $\hat{Q}_n$. Since the early stage of DNNs is unstable, we design an incremental learning to gradually increase the size of $\hat{Q}'$. We let $\tau'\!=\! \min\{\frac{n}{N_{max}}, \tau\}$, instead of $\tau$. Here $n$, $N_{max}$ are the current and maximum iterations, respectively. 

\begin{algorithm}[t]
\caption{Robust Domain Adaptation Algorithm.}\label{algorithm}
\SetKwData{Shuffle}{shuffle}\SetKwData{This}{this}\SetKwData{Up}{up}
\SetKwFunction{Union}{Union}\SetKwFunction{FindCompress}{FindCompress}
\SetKwInOut{Input}{input}\SetKwInOut{Output}{output}
\Input{ parameters $\theta_{\psi}$, $\theta_{f}$, $\theta_{f'}$, learning rate $\eta$, fixed $p_k$, fixed $\tau$, max epoch $T$, max iteration $N_{max}$}
\Output{ $\theta_{\psi}$, $\theta_{f}$, $\theta_{f'}$}
/* stage 1: filter out label-noise examples */ \\
\textbf{initialize} parameters $\theta_{f}$ \\
\For{t = 1, 2, $\dots$, $T$}{
    \textbf{collect} loss values $[\ell_{i}^t]_{i=1}^m$ of source examples $\hat{Q}_n$  \\ 
    \textbf{update} $\theta_{f} = \theta_{f} - \eta \nabla \ell(f, \psi(\hat{Q}_n))$
}
\textbf{average} each example' loss value $\bar{\ell}_{i} = \frac{1}{T}\sum_{t=1}^T\ell_{i}^t$ \\
\textbf{rank} average loss values by class in ascending order \\
\textbf{assign} $w\!=\!1$ to the top $m_k\times p_k$ examples of class $k$\\
\textbf{obtain} trusted data $\hat{Q} (\bm{w}\!=\!1)$, untrusted $\hat{Q}''(\bm{w}\!=\!0)$\\
/* stage 2: conduct domain adaptation */ \\
\textbf{initialize} parameters $\theta_{\psi}$, $\theta_{f}$, $\theta_{f'}$, $\tau'=1$ \\
\For{n = 1, 2, $\dots$, $N_{max}$}{
    \textbf{fetch} batch $\bar{Q}$ from $\hat{Q}$, $\bar{P}$ from $\hat{P}$ \\
    \textbf{update} $\theta_{f} = \theta_{f} - \eta \nabla \ell(f, \psi(\bar{Q}'))$ \\
    \textbf{update} $\theta_{\psi} = \theta_{\psi} - \eta \nabla \ell(f, \psi(\bar{Q}))$ \\
    \textbf{obtain} $\bar{Q}' = \argmin_{\bar{Q}': |\bar{Q}'| \ge \tau'|\bar{Q}|} \ell(f, \psi(\bar{Q})) $ \\
    \textbf{update} $\theta_{f'} = \theta_{f'} - \eta \nabla d_{\Lambda}^{(\rho)}(\psi(\bar{Q}'), \psi(\bar{P})) $ \\
    \textbf{update} $\theta_{\psi} = \theta_{\psi} - \eta \nabla d_{\Lambda}^{(\rho)}(\psi(\bar{Q}'), \psi(\bar{P})) $ \\
    \textbf{update} $\tau'= \min\{\frac{n}{N_{max}}, \tau\}$
}
\end{algorithm}


\begin{table*}[ht]
    \centering
    \scalebox{0.60}{
    \begin{tabular}{c|ccccccc|ccccccc|ccccccc}
         \toprule
         \multirow{2}{*}{Method} & \multicolumn{7}{c|}{Label Corruption} & \multicolumn{7}{c|}{Feature Corruption} & \multicolumn{7}{c}{Mixed Corruption}\\
         \cmidrule{2-22}
         &A$\rightarrow$W & W$\rightarrow$A & A$\rightarrow$D & D$\rightarrow$A & W$\rightarrow$D & D$\rightarrow$W & Avg &
         A$\rightarrow$W & W$\rightarrow$A & A$\rightarrow$D & D$\rightarrow$A & W$\rightarrow$D & D$\rightarrow$W & Avg &
         A$\rightarrow$W & W$\rightarrow$A & A$\rightarrow$D & D$\rightarrow$A & W$\rightarrow$D & D$\rightarrow$W & Avg\\
         \midrule
         ResNet & 47.2 & 33.0 & 47.1 & 31.0 & 68.0 & 58.8 & 47.5 & 70.2 & 55.1 & 73.0 & 55.0 & 94.5 & 87.2 & 72.5 & 58.8 & 39.1 & 69.3 & 37.7 & 75.2 & 75.5 & 59.3\\
         SPL & 72.6 & 50.0 & 75.3 & 38.9 & 83.3 & 64.6 & 64.1 & 75.8 & 59.7 & 75.7 & 56.7 & 93.9 & 87.8 & 74.9 & 77.3 & 57.5 & 78.4 & 47.5 & 93.4 & 83.5 & 72.9 \\
         MentorNet & 74.4 & 54.2 & 75.0 & 43.2 & 85.9 & 70.6 & 67.2 & 76.0 & 60.3 & 75.5 & 59.1 & 93.4 & 89.9 & 75.7 & 76.8 & 59.5 & 78.2 & 52.3 & 94.4 & 89.0 & 75.0\\
         DAN & 63.2 & 39.0 & 58.0 & 36.7 & 71.6 & 61.6 & 55.0 & 73.9 & 60.2 & 72.2 & 59.6 & 92.5 & 88.0 & 74.4 & 64.4 & 45.1 & 71.2 & 44.7 & 79.3 & 78.3 & 63.8\\
         RTN & 64.6 & 56.2 & 76.1 & 49.0 & 82.7 & 71.7 & 66.7 & 81.0
& 64.6 & 81.3 & 62.3 & 95.2 & 91.0 & 79.2 & 76.7 & 56.9 & 84.1 & 56.4 & 93.0 & 86.7 & 75.6\\
         DANN & 61.2 & 46.2 & 57.4 & 42.4 & 74.5 & 62.0 & 57.3 & 71.3 & 54.1 & 69.0 & 54.1 & 84.5 & 84.6 & 69.6 & 69.7 & 50.0 & 69.5 & 49.1 & 80.1 & 79.7 & 66.4\\
         ADDA & 61.5 & 49.2 & 61.2 & 45.5 & 74.7 & 65.1 & 59.5 & 76.8 & 62.0 & 79.8 & 60.1 & 93.7 & 89.3 & 77.0 & 69.7 & 54.5 & 72.4 & 56.0 & 87.5 & 85.5 & 70.9\\
         MDD & 74.7 & 55.1& 76.7 & 54.3 & 89.2 & 81.6 & 71.9 & 92.9 & 66.8 & 88.0 & 70.9 & \textbf{99.8} & 96.6 & 85.8 & 88.7 & 63.1 & 81.9 & 68.5 & 94.6 & 89.3 & 81.0 \\
         TCL & 82.0 & 65.7 & 83.3 & 60.5 & 90.8 & 77.2 & 76.6 & 84.9 & 62.3 & 83.7 & 64.0 & 93.4 & 91.3 & 79.9 & 87.4 & 64.6 & 83.1 & 62.2 & \textbf{99.0} & 92.7 & 81.5\\
         \midrule
         \textbf{Ours} & \textbf{89.7} & \textbf{67.2} & \textbf{92.0} & \textbf{65.5} & \textbf{96.0} & \textbf{92.7} & \textbf{83.6} & \textbf{95.1} & \textbf{68.4} & \textbf{89.4} & \textbf{72.4} & \textbf{99.8} & \textbf{97.8} & \textbf{87.2} & \textbf{93.1} & \textbf{69.5} & \textbf{92.0} & \textbf{71.5} & \textbf{99.0} & \textbf{93.1} & \textbf{86.4}\\
         \bottomrule
    \end{tabular}}
    \caption{Accuracy (\%) on \textbf{Office-31} with 40\% corruption of Label, Feature, and Both.}
    \label{tab:office-31}
\end{table*}

\begin{table*}[ht]
    \centering
    \scalebox{0.75}{
    \begin{tabular}{c|c|ccccccccccccc}
         \toprule
         \multirow{2}{*}{Method} & \multicolumn{1}{c|}{Bing-Caltech} & \multicolumn{13}{c}{Office-Home}\\
         \cmidrule{2-15}
         & B$\rightarrow$C &Ar$\rightarrow$Cl & Ar$\rightarrow$Pr & Ar$\rightarrow$Rw & Cl$\rightarrow$Ar & Cl$\rightarrow$Pr & Cl$\rightarrow$Rw  & Pr$\rightarrow$Ar & Pr$\rightarrow$Cl & Pr$\rightarrow$Rw & Rw$\rightarrow$Ar & Rw$\rightarrow$Cl &
         Rw$\rightarrow$Pr & Avg \\
         \midrule
         ResNet & 74.4 & 27.1 & 50.7 & 61.7 & 41.1 & 53.8 & 56.3 & 40.9 & 28.0 & 61.8 & 51.3 & 33.0 & 65.9 & 47.6 \\
         SPL & 75.3 & 32.4 & 56.0 & 67.4 & 41.9 & 55.3 & 57.2 & 47.9 & 32.9 & 69.3 & 60.0 & 36.2 & 70.4 & 52.2  \\
         MentorNet & 75.6 & 34.5 & 57.1 & 66.7 & 43.3 & 56.1 & 57.6 & 48.5 & 34.0 & 70.2 & 59.8 & 37.2 & 70.4 & 53.0 \\
         DAN & 75.0 & 31.2 & 52.3 & 61.2 & 41.2 & 53.1 & 54.6 & 40.7 & 30.3 & 61.5 & 51.7 & 36.7 & 67.4 & 48.5 \\
         RTN & 75.8 & 29.3 & 57.8 & 66.3 & 44.0 & 58.6 & 58.3 & 46.0 & 30.1 & 67.5 & 56.3 & 32.2 & 69.9 & 51.4 \\
         DANN & 72.3 & 32.9 & 50.6 & 60.1 & 38.6 & 49.2 & 50.6 & 39.9 & 32.6 & 60.4 & 50.5 & 38.4 & 67.4 & 47.6 \\
         ADDA & 74.7 & 32.6 & 52.0 & 60.6 & 42.6 & 53.5 & 54.3 & 43.0 & 31.6 & 63.1 & 52.7 & 37.7 & 67.5 & 49.3 \\
         MDD & 78.9  & 44.6 & 62.4 & 68.8 & 46.7 & 58.9 & 60.8  & 45.5 & 39.5 & 65.2 & 59.8 & 47.1 & 72.9  & 56.0 \\
         TCL  & 79.0 & 38.8 & 62.1 & 69.4 & 46.5 & 58.5 & 59.8 & 51.3 & 39.9 & 72.3 & 63.4 & 43.5 & 74.0 & 56.6\\
         \midrule
         \textbf{Ours} & \textbf{81.7}  & \textbf{50.8} & \textbf{68.7} & \textbf{72.3} & \textbf{55.6} & \textbf{67.4} & \textbf{67.9}  & \textbf{57.8} & \textbf{50.5} & \textbf{74.6} & \textbf{69.5} & \textbf{57.7} & \textbf{80.2}  & \textbf{64.4} \\ 
         \bottomrule
    \end{tabular}}
    \caption{Accuracy (\%) on real dataset \textbf{Bing-Caltech} and \textbf{Office-Home} with 40\% Mixed Corruption.}
    \label{tab:office-home}
\end{table*}

\begin{table}[ht]
    \centering
    \scalebox{0.75}{
    \begin{tabular}{c|cccccc|c}
         \toprule
         \multirow{2}{*}{Method} & \multicolumn{7}{c}{Office-31 40\% Mixed Corruption} \\
         \cmidrule{2-8}
         &A$\rightarrow$W & W$\rightarrow$A & A$\rightarrow$D & D$\rightarrow$A & W$\rightarrow$D & D$\rightarrow$W & Avg \\
         \midrule
         del-OCL    & 90.4 & 62.1 & 84.3 & 68.1 & 96.2 & 89.7 & 81.8\\
         del-PMD   & 92.2 & 68.8 & 89.2 & 69.7 & 98.8 & 92.7 & 85.2\\
         \textbf{Ours}       & \textbf{93.1} & \textbf{69.5} & \textbf{92.0} & \textbf{71.5} & \textbf{99.0} & \textbf{93.1} & \textbf{86.4}\\ 
         \bottomrule
    \end{tabular}}
    \caption{Ablation study by deleting OCL or PMD.}
    \label{tab:ablation study}
\end{table}

\begin{table}[ht]
    \centering
    \scalebox{0.75}{
    \begin{tabular}{c|cccccc|c}
         \toprule
         \multirow{2}{*}{Method} & \multicolumn{7}{c}{Office-31 40\% Label Corruption} \\
         \cmidrule{2-8}
         &A$\rightarrow$W & W$\rightarrow$A & A$\rightarrow$D & D$\rightarrow$A & W$\rightarrow$D & D$\rightarrow$W & Avg \\
         \midrule
         
         Online     & 83.1 & 55.4 & 79.1 & 49.9 & 88.6 & 83.8 & 73.3\\
         Offline    & \textbf{90.1} & 59.8 & 91.3 & 52.7 & 85.7 & 75.3 & 75.9\\
         \textbf{Ours}       & 89.7 & \textbf{67.2} & \textbf{92.0} & \textbf{65.5} & \textbf{96.0} & \textbf{92.7} & \textbf{83.6}\\ 
         \bottomrule
    \end{tabular}}
    \caption{An analysis of our offline curriculum learning.}
    \label{tab:loss function analysis}
\end{table}

\begin{table}[ht]
    \centering
    \scalebox{0.75}{
    \begin{tabular}{c|cccccc|c}
         \toprule
         \multirow{2}{*}{Method} & \multicolumn{7}{c}{Office-31 40\% Feature Corruption} \\
         \cmidrule{2-8}
         &A$\rightarrow$W & W$\rightarrow$A & A$\rightarrow$D & D$\rightarrow$A & W$\rightarrow$D & D$\rightarrow$W & Avg \\
         \midrule
         TCL        & 84.9 & 62.3 & 83.7 & 64.0 & 93.4 & 91.3 & 79.9 \\
         Ours-del     & 90.1 & \textbf{71.5} & \textbf{89.9} & 69.5 & 98.6 & 92.7 & 85.4\\
         Ours-res    & \textbf{95.1}  & 68.4 & 89.4 & \textbf{72.4} & \textbf{99.8} & \textbf{97.8} & \textbf{87.2} \\
         \bottomrule
    \end{tabular}}
    \caption{An analysis of the feature-noise effect on source risk.}
    \label{tab:reserve and delete analysis}
\end{table}

\begin{table}[ht]
    \centering
    \scalebox{0.75}{
    \begin{tabular}{c|ccccc|c}
         \toprule
         \multirow{2}{*}{Method} & \multicolumn{6}{c}{Wider Feature Noise Levels on A$\rightarrow$W}  \\
         \cmidrule{2-7}
         &0\% & 20\% & 40\% & 60\% & 80\% & Avg \\
         \midrule
         DANN      & 85.5  & 73.7 & 71.3 & 60.4 & 53.2 & 57.4  \\
         TCL       & 87.5  & 85.9 & 84.9 & 61.9 & 31.2 & 58.6  \\
         \textbf{Ours}      & \textbf{95.9}  & \textbf{95.7} & \textbf{95.1} & \textbf{87.2} & \textbf{80.4} & \textbf{75.7}  \\ 
         \bottomrule
    \end{tabular}}
    \caption{An analysis of different discrepancies on feature noises.}
    \label{tab:discrepancies analysis}
\end{table}

Meanwhile, since the optimization of PMD is a min-max game, we let the representation function $\psi$ to \textit{minimize} PMD and the classifier $f'$ to \textit{maximize} it. To avoid the problem of vanishing gradients, on the target domain, we use a modified cross-entropy loss: $disp_P(f', f) = \log [1\!-\!\sigma_{f(\psi(x))}(f'(\psi(x))]$. The final optimization of PMD is by
\begin{equation}
    \begin{split}
    &\min_{\psi} \max_{f' \in \mathcal{F}} \alpha \mathbb{E}_{\tilde{x}\sim \psi(\hat{Q}_n)} \log[ \sigma_{f(\psi(\tilde{x}))}(f'(\psi(\tilde{x})))] \\
    &\quad \quad \quad \; \; - \mathbb{E}_{x \sim \psi(\hat{Q}')} \log[\sigma_{f(\psi(x))}(f'(\psi(x)))] \,,
    \end{split}
\end{equation}
and,
\begin{equation}
    \begin{split}
    &\min_{\psi} \max_{f' \in \mathcal{F}} \alpha \mathbb{E}_{x\sim \psi(\hat{Q}')} \log[ \sigma_{f(\psi(x))}(f'(\psi(x)))] \\
    &\quad \quad \quad \; \; + \mathbb{E}_{x \sim \psi(\hat{P})} \log[1 - \sigma_{f(\psi(x))}(f'(\psi(x)))]\,,
    \end{split}
\end{equation}
where $\alpha$ $\triangleq$ exp $\rho$ is designed to attain the margin $\rho$.

In summary, we give a novel perspective when facing feature noises in domain adaptation. Briefly speaking, PMD injects a proxy distribution to bridge noisy source distribution and target distribution, successfully alleviating the negative impact of feature noise. The theory-induced adversarial network performs proxy distribution search and disentangles the min-max game simultaneously, successfully learning domain-invariant representation in noise environments. The optimization process uses incremental learning to control the size of proxy distribution, providing a stable solution for PMD. 

\subsection{Robust Joint Optimization}
Finally, we unify the learning with the robust conditional-weighted empirical risk ($\mathcal{E}_{\bm{w}}$) and proxy margin discrepancy in a joint min-max problem. We state the final problem as
\begin{equation}
    \begin{split}
        &\min_{f,\psi} \mathcal{E}_{\bm{w}}(\psi(\hat{Q}_n)) + \beta d_{\Lambda}^{(\rho)}(\psi(\hat{Q}_n), \psi(\hat{P}))\,, \\
        &\max_{f'} d_{\Lambda}^{(\rho)}(\psi(\hat{Q}_n), \psi(\hat{P}))\,,
    \end{split}
    \label{joint min-max}
\end{equation}
where $\beta$ is the trade-off coefficient between source error and discrepancy. In practice, we split the optimization process of Eq.~\eqref{joint min-max} into two stages: filter out label-noise examples and conduct domain adaptation, as shown in Algorithm~\ref{algorithm}.

    \begin{figure*}
        \centering
        \resizebox{1\textwidth}{!}{
        \subfigure[Label Noise]{\includegraphics[width=0.33\textwidth]{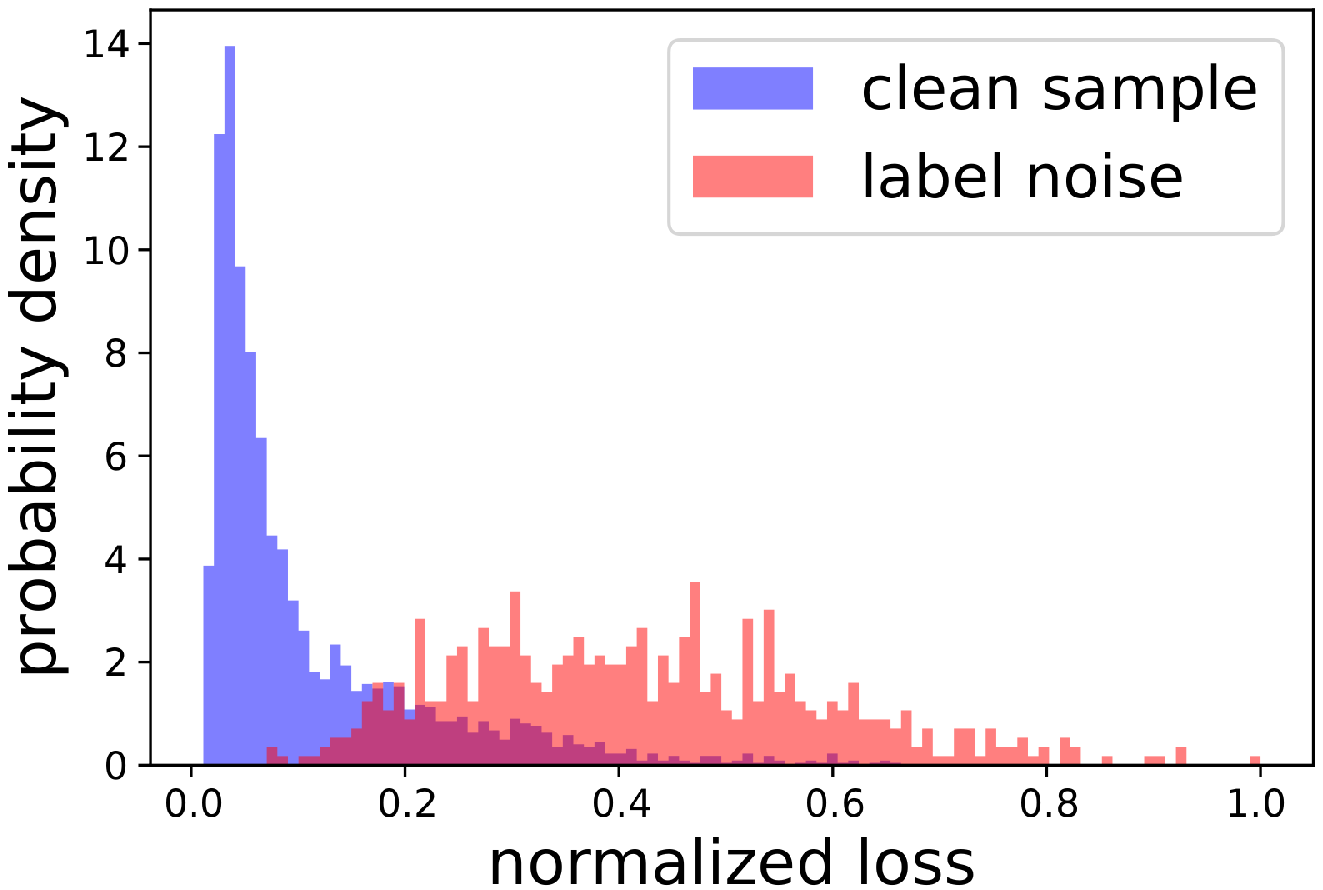}}
        \subfigure[Feature Noise]{\includegraphics[width=0.33\textwidth]{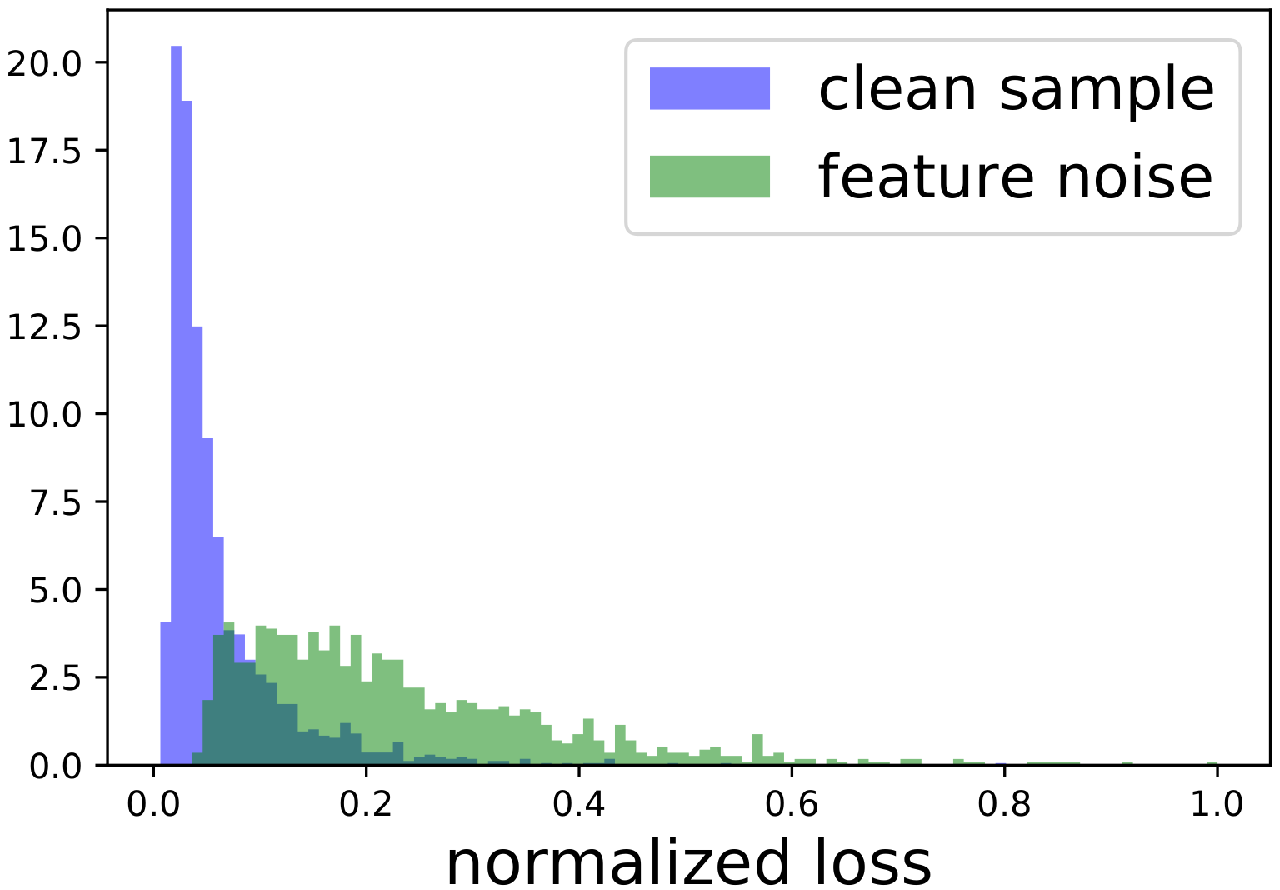}}
        \subfigure[Mixed Noise]{\includegraphics[width=0.33\textwidth]{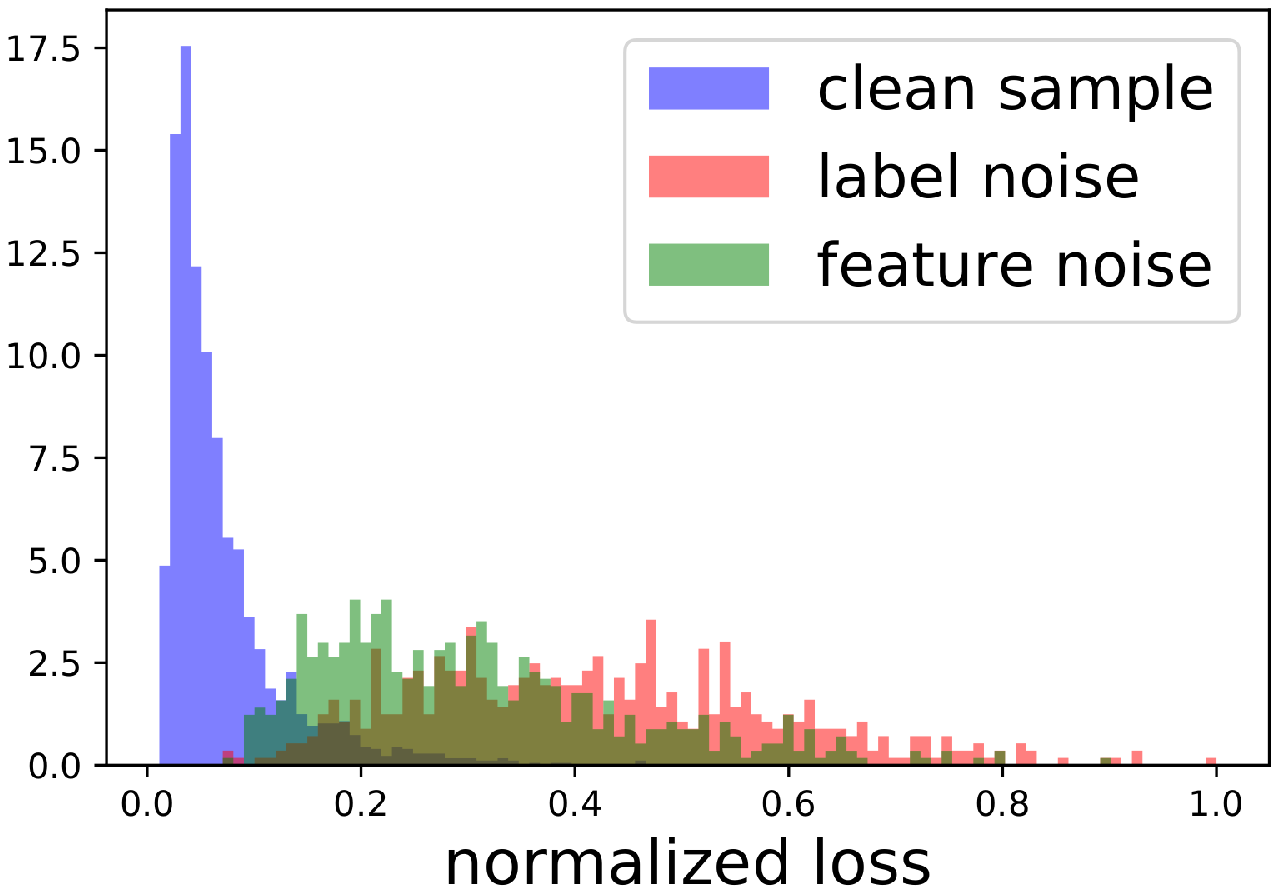}}}
    \caption{An illustration of loss distributions of the three types noises.}
    \label{loss distribution}
    \end{figure*}

    \begin{figure}
        \centering
        \resizebox{0.48\textwidth}{!}{
        \begin{tikzpicture}
            \begin{axis}[axis equal, title={},
            xlabel={Noise Level}, ylabel={Accuracy},
            axis equal=false,
            legend pos=outer north east, 
            ymin=0.4, ymax=1.0, clip=false,
            ytick={0.4,0.5,0.6,0.7,0.8, 0.9, 1.0}
            ]
            \addplot+[smooth, red, solid, very thick, mark=triangle] plot coordinates { (0,0.951) (0.2,0.936) (0.4,0.931) (0.6,0.892) (0.8,0.852)};
                \addlegendentry{Ours}
                
                \addplot+[smooth, blue, solid, thick, mark=+] plot coordinates { (0,0.875) (0.2,0.865) (0.4,0.850) (0.6,0.820) (0.8,0.78)};
                \addlegendentry{TCL}
                
                \addplot+[smooth,brown, solid, thick, mark=x] plot coordinates { (0,0.939) (0.2,0.904) (0.4,0.887) (0.6,0.848) (0.8,0.796)};
                \addlegendentry{MDD}

                \addplot+[smooth, green, solid, thick, mark=*] plot coordinates { (0,0.825) (0.2,0.758) (0.4,0.687) (0.6,0.641) (0.8,0.588)};
                \addlegendentry{ADDA}
                
                \addplot+[smooth, cyan, solid, thick, mark=square] plot coordinates { (0,0.855) (0.2,0.77) (0.4,0.697) (0.6,0.599) (0.8,0.522)};
                \addlegendentry{DANN}
                
                \addplot+[smooth, magenta, solid, thick, mark=pentagon] plot coordinates { (0,0.855) (0.2,0.802) (0.4,0.767) (0.6,0.728) (0.8,0.708)};
                \addlegendentry{RTN}
                
                \addplot+[smooth, black, solid, thick, mark=diamond] plot coordinates { (0,0.825) (0.2,0.735) (0.4,0.644) (0.6,0.565) (0.8,0.486)};
                \addlegendentry{DAN}
                
                \addplot+[smooth, violet, solid, thick, mark=otimes*] plot coordinates { (0,0.760) (0.2,0.765) (0.4,0.768) (0.6,0.75) (0.8,0.73)};
                \addlegendentry{MentorNet}
                
                \addplot+[smooth, olive, solid, thick, mark=o] plot coordinates { (0,0.75) (0.2,0.75) (0.4,0.773) (0.6,0.74) (0.8,0.722)};
                \addlegendentry{SPL}
                
                \addplot+[smooth, purple, solid, thick, mark=diamond*] plot coordinates { (0,0.74) (0.2,0.65) (0.4,0.588) (0.6,0.52) (0.8,0.48)};
                \addlegendentry{ResNet}
                \end{axis}
        \end{tikzpicture}}
        \caption{An analysis of ten methods in various mixed noise levels.}
        \label{noises levels}
        \end{figure}
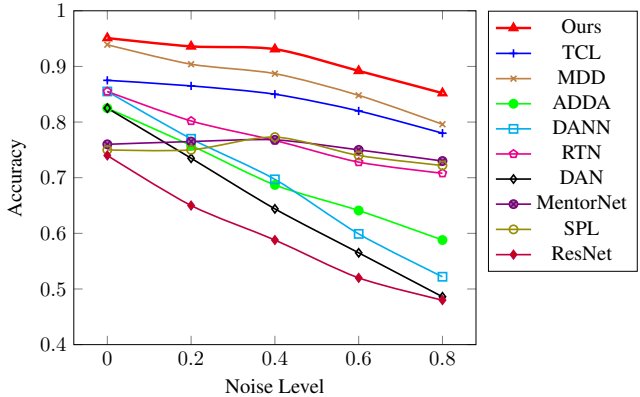

\section{Experiments}
We evaluate our algorithm on three datasets against state of the art methods. The code is at \url{https://github.com/zhyhan/RDA}.

\subsection{Setup}
\textbf{Office-Home} is a more challenging domain adaptation dataset consisting of 15,599 images with 65 unbalanced classes. It consists of four more diverse domains: \textbf{Ar}tistic images, \textbf{Cl}ip Art, \textbf{Pr}oduct images, and \textbf{R}eal-\textbf{w}orld images.

Following the protocol in the pioneering work~\cite{DBLP:conf/aaai/ShuCLW19}, we create corrupted counterparts on the above two clean datasets as follows. We make three types of corruption: label corruption, feature corruption, and mixed corruption on source domains. Label corruption uniformly changes the label of each image into a random class with probability $p_{noise}$. Feature corruption refers to each image corrupted by Gaussian blur and salt-and-pepper noise with probability $p_{noise}$. Mixed corruption refers to each image corrupted by label corruption and feature corruption with probability $p_{noise}/2$ independently. While past literature mainly studied label corruption, we also study mixed corruption with distribution shifts.

\textbf{Bing-Caltech} is a real noisy dataset created with \textbf{B}ing and \textbf{C}altech-256 datasets with 256 classes. The Bing dataset consists of rich mixed noises because it was collected by retrieving from the Bing image search engine using the category labels of Caltech-256. We naturally set Bing as a source domain, while Caltech-256 as the target domain.

We compare our designed Robust Domain Adaptation (\textbf{RDA}) algorithm with state of the art methods: \textbf{ResNet-50}~\cite{DBLP:conf/cvpr/HeZRS16}, Self-Paced Learning~(\textbf{SPL})~\cite{DBLP:conf/nips/KumarPK10}, \textbf{MentorNet}~\cite{DBLP:conf/icml/JiangZLLF18}, Deep Adaptation Network~(\textbf{DAN})~\cite{DBLP:conf/icml/LongC0J15}, Residual Transfer Network~(\textbf{RTN})~\cite{DBLP:conf/nips/LongZ0J16}, Domain Adversarial Network~(\textbf{DANN})~\cite{DBLP:conf/icml/GaninL15}, Adversarial Discriminative Domain Adaptation~(\textbf{ADDA})~\cite{tzeng2017adversarial}, Margin Disparity Discrepancy based algorithm~(\textbf{MDD})~\cite{DBLP:conf/icml/0002LLJ19}, and Transferable Curriculum Learning~(\textbf{TCL})~\cite{DBLP:conf/aaai/ShuCLW19}. Note that \textbf{TCL} is the pioneering work with the same setup of datasets as ours.

We implement our algorithm in Pytorch. We use \textbf{ResNet-50} as the representation function with parameters pre-trained from ImageNet. The main classifier and adversarial classifier are both 2-layer neural networks. We set the early training epoch $T$ to 30. $\alpha$ is set to 3 and $\beta$ is set to 0.1 according to~\cite{DBLP:conf/icml/0002LLJ19}. We use mini-batch SGD with the Nesterov momentum 0.9. The initial learning rate of the classifiers $f$ and $f'$ is 0.001, which is ten times than of the representation function $\psi$. The set of $r_k$ and $\tau$ depends on noise rates.

\subsection{Results}
Table~\ref{tab:office-31} reports the results on Office-31 under 40\% corruption of label, feature, and both. Our algorithm significantly outperforms existing domain adaptation methods and noise label methods (MentorNet, SPL) on almost all tasks. Note that the standard domain adaptation methods (DANN, MDD, etc.) suffer from over-fitting under noises while our algorithm performs stably positive adaptation, which demonstrates its robustness and generalization ability. While the pioneering work TCL uses additional entropy minimization to enhance its performance, our algorithm outperforms it by a large margin. Table~\ref{tab:office-home} reports the results on the real dataset Bing-Caltech and the synthetic dataset Office-Home under 40\% mixed corruption, where we make a remarkable performance boost. 


\begin{figure*}[ht]
    \centering
    \resizebox{1\textwidth}{!}{
        \subfigure[DANN]{\includegraphics[width=0.33\textwidth]{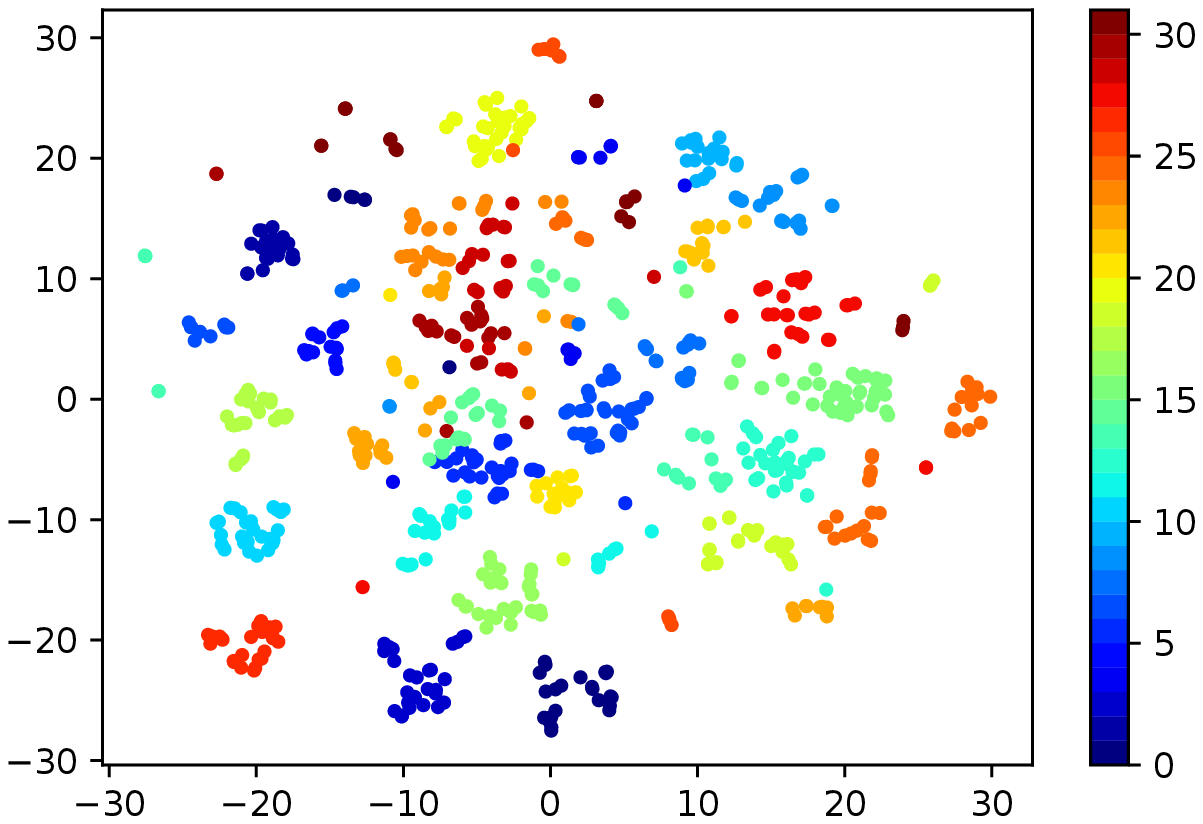}}
        \subfigure[TCL]{\includegraphics[width=0.33\textwidth]{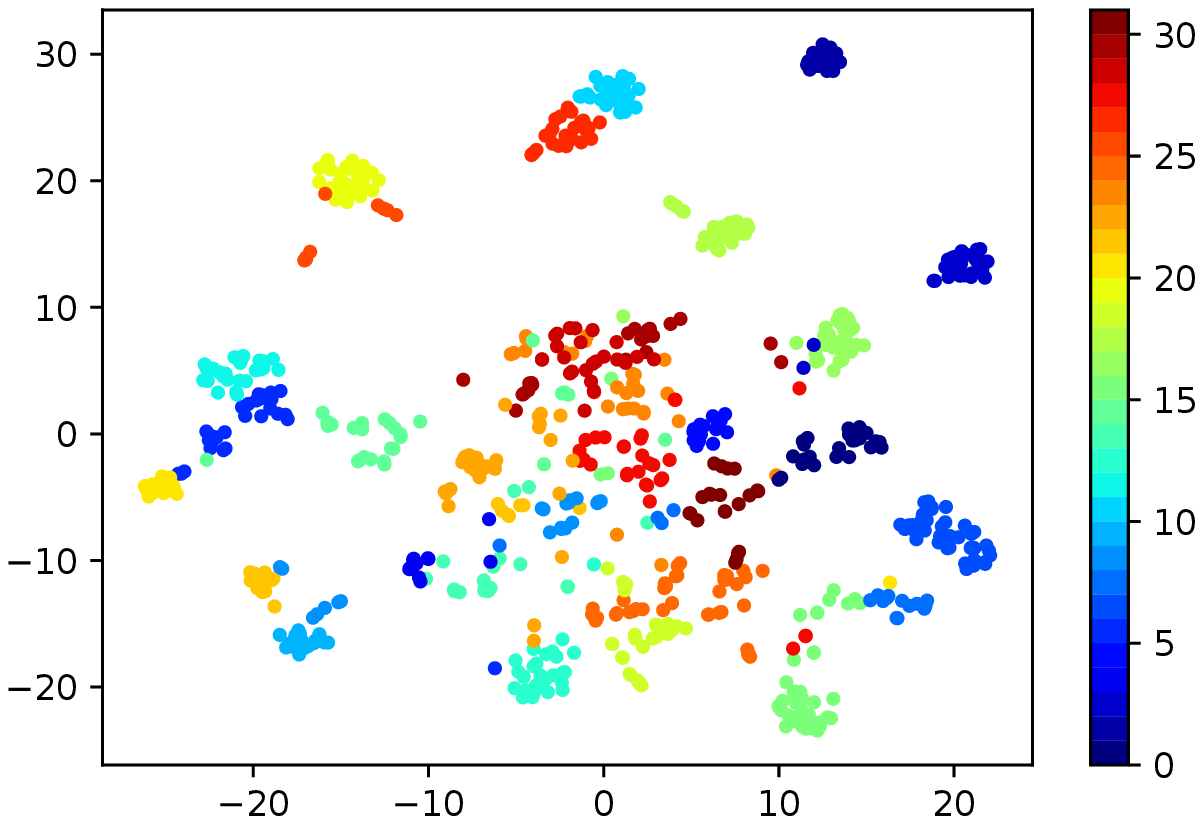}}
        \subfigure[Ours]{\includegraphics[width=0.33\textwidth]{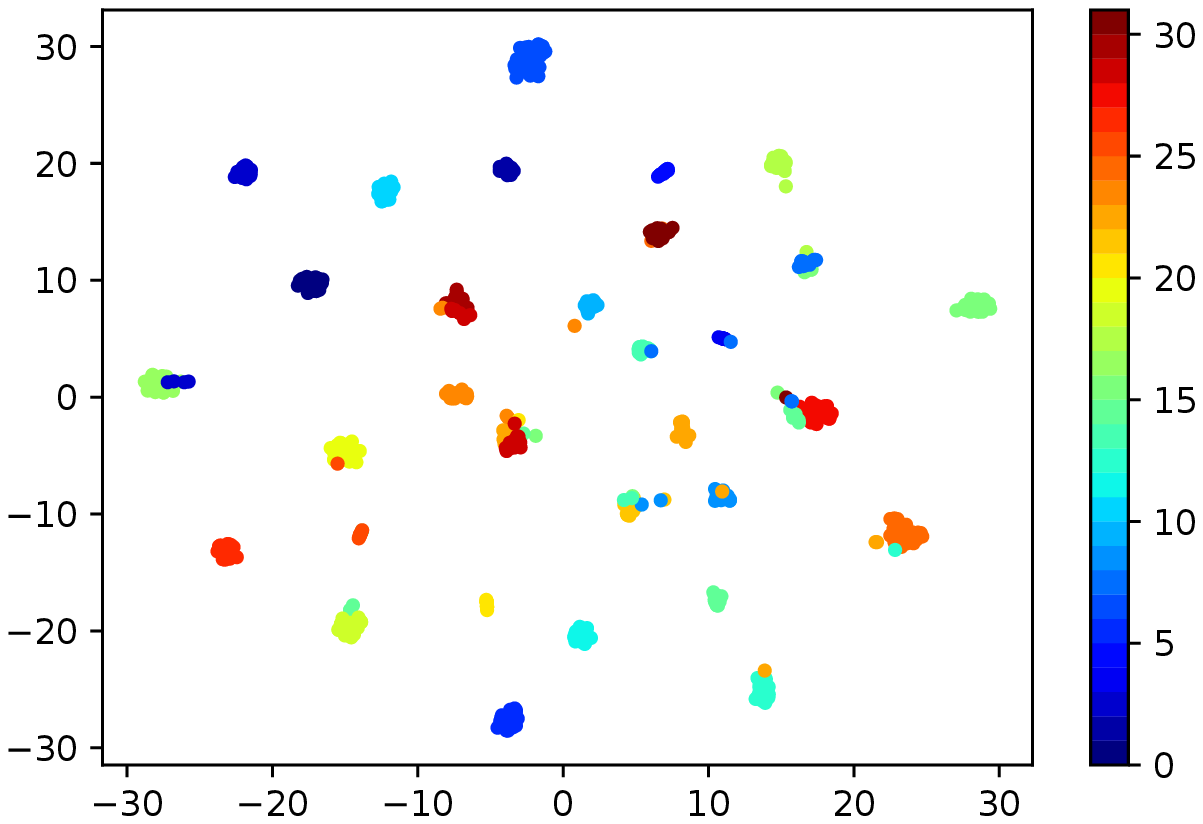}}}
    \caption{The t-SNE visualization of class features on target domain.}
    \label{t-SNE}
\end{figure*}

\subsection{Analysis}
\paragraph{Ablation Study.}
Table~\ref{tab:ablation study} reports the results of our ablation study. Abandoning either the Offline Curriculum Learning (\textbf{del-OCL}) or the Proxy Margin Discrepancy (\textbf{del-PMD}) results in a significant decline, which demonstrates the efficacy and robustness of both methods in noisy environments.

\paragraph{Offline Curriculum Learning.}
We dissect the strengths of offline curriculum learning. Table~\ref{tab:loss function analysis} reports the results of three curriculum modes: online, offline, ours (Offline Curriculum Learning) on Office-31 under 40\% label corruption. The online mode performs worse than the offline mode, which performs worse than us, proving the necessity of considering the averaged loss and class priors. Furthermore, we also justify our claim that we need to treat label-noise and feature-noise examples separately. As shown in Table~\ref{tab:reserve and delete analysis}, our method gets better performance than TCL that treats them indiscriminately. Table~\ref{tab:reserve and delete analysis} also reports that reserving feature-noise examples (\textbf{ours-res}) generally gets better results than deleting them (\textbf{ours-del} and \textbf{TCL}) when optimizing empirical source risk.

\paragraph{Proxy Margin Discrepancy.}
While Table~\ref{tab:office-31} has demonstrated the strengths of our proxy margin discrepancy of mitigating the negative influence from feature-noise examples, we provide a broader spectrum for more in-depth analysis. Table~\ref{tab:discrepancies analysis} reports the results of representative methods on wider feature-noise levels. Our method maintains stable performance while DANN and TCL drop rapidly with the increase of noise levels.

\paragraph{Noise Levels.}
Fig.~\ref{noises levels} reports the results in various mixed noise levels on A$\rightarrow$W task. Our method outperforms all the compared methods at each level, which demonstrates that our method is more robust under severe noisy environments. In particular, our methods achieve the best result when the noise level is 0\%, which proves that our method can also fit into the standard domain adaptation scenarios.


\paragraph{Loss Distribution.}
Fig.~\ref{loss distribution} exhibits the loss distributions of the three types of noises of 40\% on the Office-31 Amazon dataset. Fig.~\ref{loss distribution}~(a,~b) verifies that correct examples have smaller losses than incorrect examples. Fig.~\ref{loss distribution}~(c) verifies that feature-noise examples generally have smaller losses than label-noise examples, which fits our intuition that label noise is more harmful than feature noise. 

\paragraph{Feature Visualization.}
Fig.~\ref{t-SNE} illustrates the t-SNE embeddings~\cite{DBLP:conf/icml/DonahueJVHZTD14} of the learned representations by DANN, TCL, and ours, on 40\% mixed corruption of A$\rightarrow$W task. While the learned features of DANN, and TCL are mixed up, ours is more discriminative in every class on the target domain, which verifies that our method can learn domain-invariant representations in noise environments.
\section{Conclusion}
We presented a new analysis of domain adaptation in noisy environments, an under-explored but more realistic scenario. We also proposed an offline curriculum learning and a new proxy discrepancy tailored to label and feature noises, respectively. Our methods are more robust for achieving real-world applications in noisy non-stationary environments. The theory-induced algorithm yields the state of the art results in all tasks. 

\section*{Acknowledgements}
This work is supported by the National Natural Science Foundation of China (61876098, 61573219, 61701281), the National Key R\&D Program of China (2018YFC0830100, 2018YFC0830102), and the Fostering Project of Dominant Discipline and Talent Team of Shandong Province Higher Education Institutions.


%

\bibliographystyle{named}
\bibliography{ijcai20}

\end{document}